\pdfoutput=1

\documentclass[11pt]{article}

\usepackage[]{emnlp2021} 

\usepackage{times}
\usepackage{latexsym}

\usepackage[T1]{fontenc}

\usepackage[utf8]{inputenc}

\usepackage{microtype}

\usepackage{booktabs}
\usepackage{microtype}
\usepackage{adjustbox}
\usepackage{tabularx}
\usepackage{makecell}
\usepackage{multirow}
\usepackage{enumitem}
\usepackage{caption}
\usepackage{bbding}
\definecolor{mybrown}{RGB}{189, 96, 39}

\title{Semantic Categorization of Social Knowledge for\\ Commonsense Question Answering}

 \author{Gengyu Wang\textsuperscript{1} ~~ Xiaochen Hou\textsuperscript{2} \\ {\bf Diyi Yang\textsuperscript{3} ~~ Kathleen McKeown\textsuperscript{1}  ~~ Jing Huang\textsuperscript{2} }\\
\textsuperscript{1}Columbia University, New York, NY \\ 
\textsuperscript{2}JD AI Research, Mountain View, CA \\ 
\textsuperscript{3}Georgia Institute of Technology, Atlanta, GA \\
 \texttt{gengyu.wang@columbia.edu, houxiaochen1994@hotmail.com}\\  \texttt{diyi.yang@cc.gatech.edu, kathy@cs.columbia.edu, jing.huang@jd.com}}

\begin{document}
\maketitle
\begin{abstract}
Large pre-trained language models (PLMs) have led to great success on various commonsense question answering (QA) tasks in an end-to-end fashion. However, little attention has been paid to what commonsense knowledge is needed to deeply characterize these QA tasks. In this work, we proposed to categorize the semantics needed for these tasks using the SocialIQA as an example. Building upon our labeled social knowledge categories dataset on top of SocialIQA, we further train neural QA models to incorporate such social knowledge categories and relation information from a knowledge base. Unlike previous work, we observe our models with semantic categorizations of social knowledge can achieve comparable performance with a relatively simple model and smaller size compared to other complex approaches.
\end{abstract}

\section{Introduction} \label{introduction}

Recently, large pre-trained language models (PLMs) \cite{devlin2018bert,raffel2019exploring,liu2019roberta} have been widely used on various commonsense QA tasks such as CommonsenseQA ~\cite{Malaviya2019}, SocialIQA~\cite{sap2019socialiqa}, and \citet{mostafazadeh2016corpus, huang2019cosmos, Boratko2020,Levesque2011,roemmele2011choice}.
One line of work~\cite{khashabi2020unifiedqa} improved the performances of these QA tasks by aggregating more QA data and using even bigger PLM T5~\cite{raffel2019exploring}. 
Other line of work tried to supplement the question context with retrieval of related knowledge from external knowledge bases (KB), or re-trained PLMs under the guidance of KBs~\cite{shen2020exploiting, shwartz2020unsupervised, mitra2019additional, ji2020generating, ji2020language}.


\begin{figure}[h!] 
\includegraphics[scale=0.67]{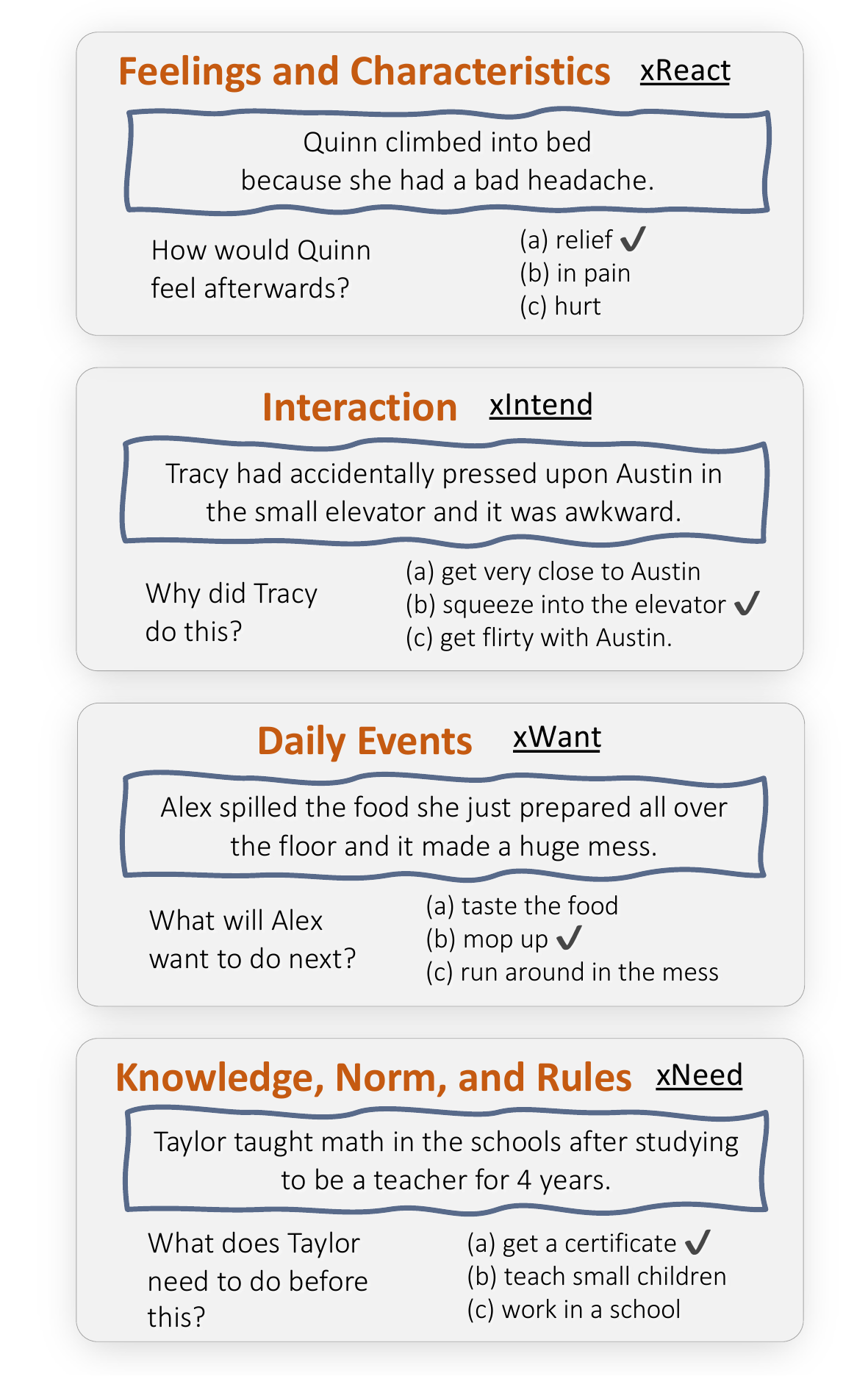}
\centering
\caption{SocialIQA Examples for \textbf{\textcolor{mybrown}{Social Knowledge Categories}} and \textbf{\underline{Question Relation Type}}}
\label{figure:examepls}
\end{figure}

However, very little past research has paid attention to the specific question/context knowledge types that are needed for these commonsense QA tasks. Therefore, in this paper, we go deeper into the QA task context and take a closer look at the semantics on what additional information can be inferred from the given question-answer context in order to answer a question. 
Using SocialIQA as an example, we propose to add two new context types (See Figure \ref{figure:examepls}) into the neural QA model: one on question relation type derived from ATOMIC (which was used to create SocialIQA), and another knowledge category type from our own constructed social knowledge taxonomy. While the question relation type derived from ATOMIC is restricted on ATOMIC related datasets, our constructed social knowledge category type has the potential
be generally applied to other social knowledge related tasks.

To fully utilize these two new types of context information, we adopt a simple yet effective way of integrating this information to help in the neural QA model. Specifically, we concatenate each QA pair with its assigned question relation type or its social knowledge category as the input to a PLM (say RoBERTa~\cite{liu2019roberta}), and fine tune the RoBERTa model for the SocialIQA task. Our experimental results show that this simple and interpretable method not only outperforms the RoBERTa baseline model, but also achieves comparable performances as that of previous work, which adopted much more complex models to encode external knowledge or re-train large language models. 

In terms of creating efficient and sustainable models for QA tasks, our work illustrates the importance of deep understanding of what knowledge is required for the specific commonsense tasks.
Our constructed social knowledge category, along with experiment code and human annotations on some of the SocialIQA data, are released to the research community.\footnote{https://github.com/posuer/social-commonsense-knolwedge}

\section{Related Work}
\paragraph{SocialIQA Task}
Most previous works on SocialIQA task involve either large size of pre-trained models, and datasets \cite{khashabi2020unifiedqa, lourie2021unicorn} or complicated models that heavily rely on external knowledge bases \citep{shen2020exploiting, shwartz2020unsupervised, mitra2019additional, ji2020generating, ji2020language, chang2020incorporating}. Among them, UnifiedQA \citep{khashabi2020unifiedqa} achieved impressive performance by fine-tuning 11B T5 model \cite{raffel2019exploring} with 17 existing QA datasets. Unlike previous efforts, our work achieves comparable performance with a relatively small model and simple knowledge extraction method that does not rely on knowledge bases nor require additional pretraining. 

\paragraph{Commonsense Categorization in NLP}
\citet{LoBue2011} proposed form-based and content-based categories for commonsense knowledge that is involved in recognizing textual entailment.
\citet{Boratko} refined the categorization method for knowledge and reasoning proposed for a QA dataset ARC \cite{clark2018think}. 
The human-annotated relevant sentences was used only for improving the retrieval model, not for the ARC task.
In summary none of these work had attempted to leverage such categories in the intended task. 

\paragraph{Social Knowledge Categorization}
\citet{kiesler19831982} proposed a taxonomy for two-dimensional interpersonal behavior, which consists of 16 segments and 128 subclasses. \citet{CowenE7900} identified 27 distinct varieties of human emotion, such as anger, excitement, relief, etc. Recently, \citet{Forbes} introduced a formalism to study people's social and moral norms over daily life situations, which includes 12 different dimensions of people's judgments. 
Motivated by these prior work that covered different aspects of knowledge of daily events and social interactions, our work provides a comprehensive overview of social knowledge needed by the SocialIQA task.

\section{Methodology}
This section presents two approaches to model the underlying semantics and knowledge of the SocialIQA task, together with a simple yet effective method that leverages these knowledge types to improve QA models. 

\subsection{Question Relation Type}\label{method_detail}
The SocialIQA dataset was derived from the ATOMIC~\cite{sap2019atomic}. ATOMIC is a knowledge base that focuses on everyday social commonsense knowledge organized as ten types of if-then relations.
Based on this observation, we tag each question in SocialIQA according to its relation types in ATOMIC by conducting rule-based mapping between them. 
Specifically, we match keywords in the questions and use the Spacy model \cite{spacy} to detect subjective and objective in context sentences.~\footnote{Take the fourth instance in Figure \ref{figure:examepls} as an example; we firstly match the word ``need" in question to the "Need" relation, then detect the name ``Taylor" is subjective in the context, so we assign ``xNeed" to this question.}

Once the mapped types are obtained for each SocialIQA question, we transfer such information to QA models by simply concatenating the tags to original QA examples in the format of $[Context, \textrm{SEP}, Question, \underline{Tag}, Answer]$ as input to a PLM for fine-tuning.~\footnote{All the tags are added into the model's vocabulary as special tokens, for instance $[xNeed]$, and they are concatenated to QA examples in text form.}
Although we use RoBERTa model in this work, our method is generic and can be  applied to any PLMs.


\subsection{Social Knowledge Categorization}
Although utilizing ATOMIC to obtain relation type is straightforward and simple, it is restricted to datasets derived from ATOMIC. Inspired by previous work around emotion and social interactions in psychology \cite{kiesler19831982, CowenE7900, Forbes}, we propose a taxonomy to categorize social commonsense knowledge into types, which can be generally applied to other social commonsense reasoning related tasks. 
As shown in Figure \ref{figure:examepls}, 
this taxonomy includes four categories as follows:

\paragraph{Feelings and Characteristics} 
involves personal feelings and characteristics. Specifically, it
includes the development of subsequent 
feelings and emotions, and events triggered by personal feelings, emotions or characteristics; 
the feelings and emotions caused by a certain event; 
and the personal characteristics reflected by a particular event.

\paragraph{Interaction} includes events, daily life habits and experiences caused by interactions or emotions among two or more people, as well as interactions and possible responsibilities and obligations between individuals and groups.

\paragraph{Daily Events} deal with relationships between daily events,  habits, and life experiences. In this category, most situations only involve individuals. Even if multiple people are involved, the focus is on the event itself rather than on the interaction between people. For example, in such a scenario, ``Two people went to the hair salon together, what will they do next?", it will be classified into this category even though it involves two persons.

\paragraph{Knowledge, Norm, and Rules}
Unlike daily events, the events included in this category usually involve social or scientific knowledge and rules that are written in documents or books. The knowledge and rules here may pertain to various topics such as legislation, law, career development, social identity, and medical care.

\subsection{SocialIQA-Category Dataset}



We manually annotate 
some of the SocialIQA data with our proposed four social knowledge categories. 
Among the total 800 examples in this dataset, 600 training examples are selected from SocialIQA training data, and 200 dev examples are selected from SocialIQA dev data. 


\paragraph{Dataset Creation} Since the annotation requires fully understanding the social knowledge category, 
two listed authors annotate two rounds of 50 examples, discuss the disagreements in the middle. The percentage of agreement on the second round is higher than 95\%, which indicates that these categories are well-defined. Then each of the two annotators is responsible for another 350 examples separately. 



\paragraph{Knowledge Category Prediction}\label{method_detail2}
We fine-tune RoBERTa large model with the SocialIQA-Category training set and achieve up to 80\% accuracy on the dev set for this four-category classification task. The trained model is used to assign category labels to the whole SocialIQA dataset automatically. Follow the same method in section \ref{method_detail}, the obtained category labels are concatenated to original examples in the format of $[Context, \underline{label}, \textrm{SEP}, Question, Answer]$ to train and test the QA model. 



\begin{table}
\centering
\begin{adjustbox}{max width=0.49\textwidth}
\begin{tabular}{l c c }
\toprule
Model & Dev & Test \\

\midrule
RoBERTa large & 77.4 & 77.0 \\
 + \textbf{Question Relation} & 79.0 & 78.3\\
 + \textbf{Social Knowledge Cat}egory & 79.4 & 78.5\\
 + \textbf{Question Relation + Social Knowledge Cat} & 79.8 & 78.5 \\
\midrule

Knowledge Source~\cite{mitra2019additional} & 79.5 & 78.0 \\
GLM~\cite{shen2020exploiting} & 79.6 & 78.6 \\

\midrule
Ablation Study: &  & \\
 + Question Relation \textbf{(Random)} & 77.6 & 75.3\\
 + Social Knowledge Category \textbf{(Random)} & 77.6 & 76.4\\
\bottomrule
\end{tabular}
\end{adjustbox}
\caption{Accuracy on SocialIQA test set
}
\label{table:mainresults}
\end{table}
\begin{table*}
\centering
\begin{adjustbox}{max width=\textwidth}
\begin{tabular}{ c l l l l l l l l l l }
\toprule
Model & xIntent & xNeed & xAttr & xReact & xWant & xEffect & oReact & oWant & oEffect & Other \\
\midrule
RoBERTa-large & 0.20 & 0.22 & 0.26 & 0.24 & 0.26 & 0.24 & 0.23 & 0.23 & 0.28 & 0.21 \\
+ \textbf{Question Relation} & \textbf{0.19}$\downarrow$ & \textbf{0.19}$\downarrow$ & \textbf{0.24}$\downarrow$ & \textbf{0.23}$\downarrow$ & \textbf{0.24}$\downarrow$ & \textbf{0.21}$\downarrow$ & 0.23 & \textbf{0.20}$\downarrow$ & \textbf{0.26}$\downarrow$ & \textbf{0.13}$\downarrow$\\
+ \textbf{Question Relation + Social Knowledge Cat.} & 0.22 & 0.23 & \textbf{0.21}$\downarrow$ & \textbf{0.21}$\downarrow$ & \textbf{0.21}$\downarrow$ & 0.21 & 0.23 & \textbf{0.18}$\downarrow$  & \textbf{0.25}$\downarrow$ & 0.17 \\
\bottomrule
\end{tabular}
\end{adjustbox}
\caption{Error Rate Distribution of \textit{Question Relation} model on SocialIQA Dev Set }
\label{table:Q2RelDist}
\end{table*}
\begin{table}
\centering
\begin{adjustbox}{max width=0.5\textwidth}
\begin{tabular}{ c c c c c }
\toprule
Model & \makecell{Feelings and\\Characteristics} & Interaction & \makecell{Daily\\Events} & \makecell{Knowledge, \\ Norm, and Rules} \\
\midrule
RoBERTa-large & 0.25 & 0.21 & 0.24 & 0.21 \\
\midrule
\textbf{\makecell[c]{+Social Knowledge \\Category}} & \textbf{0.21}$\downarrow$ & \textbf{0.19}$\downarrow$ & \textbf{0.20}$\downarrow$ & 0.22\\
\midrule
\textbf{\makecell[c]{+ Question Relation \\+ Soc. Knowl. Cat.}} & 0.21 & \textbf{0.16}$\downarrow$ & 0.20 & 0.26 \\
\bottomrule
\end{tabular}
\end{adjustbox}
\caption{Error Rate Distribution of \textit{Social Knowledge Category} model on SocialIQA-Category Dev Set}
\label{table:CatogoryDist}
\end{table}
\section{Experiments and Results}
We use the SocialIQA data set for our experiments. SocialIQA contains $33,410$ training examples, $1,954$ dev examples and $2,224$ test examples.

\paragraph{Models \& Baselines}
We employ the RoBERTa-large model as baseline. Our proposed method uses RoBERTa-large models in the same way of the baseline, except concatenating the tags or labels to original QA examples (described in Section \ref{method_detail} and \ref{method_detail2}).
We also compare our methods with the following published models on SocialIQA: \textit{GLM}~\cite{shen2020exploiting} re-trains the RoBERTa model by injecting structured knowledge from the knowledge graph.  \textit{Knowledge Source}~\cite{mitra2019additional} concatenates the question, answer, and the context as the query to retrieve and re-rank the top ten sentences from ATOMIC, and then fuses them into the QA model to select the right answer.

\paragraph{Training Setup}
The hyperparameters are selected based on the best performing model on the dev set.
We use grid search to fine-tune the model, and select select  the learning  rate from $\{1e-5, 2e-5\}$, batch size from $\{4,8\}$ and gradient accumulation from $\{4,8,16\}$. The  model  is  trained  up  to 4 epochs.~\footnote{Details about the experiment setting is in Appendix.}

\paragraph{Results}
We report the accuracy results on the SocialIQA test set in Table \ref{table:mainresults}. 

\begin{itemize}[leftmargin=6pt]
\setlength{\itemsep}{0pt}
    \setlength{\parskip}{2pt}
        \item 
        Both ways of using knowledge type information outperform the RoBERTa baseline models. The paired t-test shows that \textit{Social Knowledge Category} achieves significant gains over the RoBERTa model with $p<0.05$, and \textit{Question Relation} is a little short of significance on test set ($p=0.06$).
        
        \item Compared with \textit{GLM} and \textit{Knowledge Source} that require large amounts of engineering work to explore external knowledge from ATOMIC, our simple and direct utilization of the ATOMIC relations and social knowledge categorization achieves competitive performances. 
        
        \item The naive combination of question relation type and social knowledge category shows no gains over any single model. One reason may be that these two types of relations are not entirely orthogonal to each other. 
\end{itemize}

\paragraph{Error Analysis}
We conduct detailed error analysis to examine the performance gains from \textit{Question Relation} and \textit{Social Knowledge Category} models. The results are presented in Table \ref{table:Q2RelDist} and \ref{table:CatogoryDist}.

For the \textit{Question Relation} model, we present the error rate under different relations on the socialIQA dev set. As we can see, the \textit{Question Relation} model achieves consistent improvements on almost all relations, which proves the effectiveness of incorporating the logical relation type.

We also compare the error rate of the \textit{Social Knowledge Category} and the \textit{RoBERTa-large} on the manually annotated dev set with 200 examples.
We observe performance gains on all categories except the Knowledge, Norm and Rules category. 
This suggests that current QA models still struggle with questions involving knowledge and norms, calling for more sophisticated techniques to reason over these social and scientific knowledge. 
\footnote{More examples and error analyses on these two types of relations can be found in Appendix. }


\paragraph{Ablation Study}
We conduct the ablation study for our proposed method described in Section \ref{method_detail} and \ref{method_detail2}. While training and testing, the mapped relation type tag or predicted knowledge category label is replaced by a randomly chosen tag or label. The experiment result presented in Table \ref{table:mainresults} shows that the randomly chosen tags or labels could not help the QA models. The performance gains from knowledge of Question Relation or Social Knowledge Category are indeed valid.

\section{Conclusion}
In this work, using the SocialIQA task as an example, 
we integrate two different knowledge types into the QA model training: one based on question relations, and the other is our own defined social knowledge category. 
Experimental results demonstrated that incorporating semantic categorizations of social knowledge into QA models helps boost performances on the social commonsense QA task. The proposed simply ways of incorporating knowledge into the model also achieved comparable performances to these much complicated models. 

\section{Ethics}
We create a dataset, SocialIQA-Category, by annotating part of the SocialIQA dataset \cite{sap2019atomic}. SocialIQA dataset is accessible to the public and can be downloaded from an open URL. All the annotations are done by the listed authors of this paper. 
The annotations only include the aforementioned relation type and social knowledge category. 
Our work focuses on QA tasks, specifically the SocialIQA task. Neural models that are created for this task are not supposed to solve any real-world problem. 
In terms of environmental consequences, all of our experiments are done with the RoBERTa model. Models training for SocialIQA is usually done within 1 hour on a single GPU.

\bibliography{anthology,custom}

\begin{thebibliography}{26}
\expandafter\ifx\csname natexlab\endcsname\relax\def\natexlab#1{#1}\fi

\bibitem[{Boratko et~al.(2020)Boratko, Li, O{'}Gorman, Das, Le, and
  McCallum}]{Boratko2020}
Michael Boratko, Xiang Li, Tim O{'}Gorman, Rajarshi Das, Dan Le, and Andrew
  McCallum. 2020.
\newblock \href {https://doi.org/10.18653/v1/2020.emnlp-main.85} {{P}roto{QA}:
  A question answering dataset for prototypical common-sense reasoning}.
\newblock In \emph{Proceedings of the 2020 Conference on Empirical Methods in
  Natural Language Processing (EMNLP)}, pages 1122--1136, Online. Association
  for Computational Linguistics.

\bibitem[{Boratko et~al.(2018)Boratko, Padigela, Mikkilineni, Yuvraj, Das,
  McCallum, Chang, Fokoue-Nkoutche, Kapanipathi, Mattei, Musa, Talamadupula,
  and Witbrock}]{Boratko}
Michael Boratko, Harshit Padigela, Divyendra Mikkilineni, Pritish Yuvraj,
  Rajarshi Das, Andrew McCallum, Maria Chang, Achille Fokoue-Nkoutche, Pavan
  Kapanipathi, Nicholas Mattei, Ryan Musa, Kartik Talamadupula, and Michael
  Witbrock. 2018.
\newblock \href {https://doi.org/10.18653/v1/W18-2607} {A systematic
  classification of knowledge, reasoning, and context within the {ARC}
  dataset}.
\newblock In \emph{Proceedings of the Workshop on Machine Reading for Question
  Answering}, pages 60--70, Melbourne, Australia. Association for Computational
  Linguistics.

\bibitem[{Chang et~al.(2020)Chang, Liu, Gopalakrishnan, Hedayatnia, Zhou, and
  Hakkani-Tur}]{chang2020incorporating}
Ting-Yun Chang, Yang Liu, Karthik Gopalakrishnan, Behnam Hedayatnia, Pei Zhou,
  and Dilek Hakkani-Tur. 2020.
\newblock Incorporating commonsense knowledge graph in pretrained models for
  social commonsense tasks.
\newblock In \emph{Proceedings of Deep Learning Inside Out (DeeLIO): The First
  Workshop on Knowledge Extraction and Integration for Deep Learning
  Architectures}, pages 74--79.

\bibitem[{Clark et~al.(2018)Clark, Cowhey, Etzioni, Khot, Sabharwal, Schoenick,
  and Tafjord}]{clark2018think}
Peter Clark, Isaac Cowhey, Oren Etzioni, Tushar Khot, Ashish Sabharwal, Carissa
  Schoenick, and Oyvind Tafjord. 2018.
\newblock Think you have solved question answering? try arc, the ai2 reasoning
  challenge.
\newblock \emph{arXiv preprint arXiv:1803.05457}.

\bibitem[{Cowen and Keltner(2017)}]{CowenE7900}
Alan~S. Cowen and Dacher Keltner. 2017.
\newblock \href {https://doi.org/10.1073/pnas.1702247114} {Self-report captures
  27 distinct categories of emotion bridged by continuous gradients}.
\newblock \emph{Proceedings of the National Academy of Sciences},
  114(38):E7900--E7909.

\bibitem[{Devlin et~al.(2019)Devlin, Chang, Lee, and
  Toutanova}]{devlin2018bert}
Jacob Devlin, Ming-Wei Chang, Kenton Lee, and Kristina Toutanova. 2019.
\newblock \href {https://doi.org/10.18653/v1/N19-1423} {{BERT}: Pre-training of
  deep bidirectional transformers for language understanding}.
\newblock In \emph{Proceedings of the 2019 Conference of the North {A}merican
  Chapter of the Association for Computational Linguistics: Human Language
  Technologies, Volume 1 (Long and Short Papers)}, pages 4171--4186,
  Minneapolis, Minnesota. Association for Computational Linguistics.

\bibitem[{Forbes et~al.(2020)Forbes, Hwang, Shwartz, Sap, and Choi}]{Forbes}
Maxwell Forbes, Jena~D. Hwang, Vered Shwartz, Maarten Sap, and Yejin Choi.
  2020.
\newblock \href {https://doi.org/10.18653/v1/2020.emnlp-main.48} {Social
  chemistry 101: Learning to reason about social and moral norms}.
\newblock In \emph{Proceedings of the 2020 Conference on Empirical Methods in
  Natural Language Processing (EMNLP)}, pages 653--670, Online. Association for
  Computational Linguistics.

\bibitem[{Honnibal et~al.(2020)Honnibal, Montani, Van~Landeghem, and
  Boyd}]{spacy}
Matthew Honnibal, Ines Montani, Sofie Van~Landeghem, and Adriane Boyd. 2020.
\newblock \href {https://doi.org/10.5281/zenodo.1212303} {{spaCy:
  Industrial-strength Natural Language Processing in Python}}.

\bibitem[{Huang et~al.(2019)Huang, Bras, Bhagavatula, and
  Choi}]{huang2019cosmos}
Lifu Huang, Ronan~Le Bras, Chandra Bhagavatula, and Yejin Choi. 2019.
\newblock Cosmos qa: Machine reading comprehension with contextual commonsense
  reasoning.
\newblock \emph{arXiv preprint arXiv:1909.00277}.

\bibitem[{Ji et~al.(2020{\natexlab{a}})Ji, Ke, Huang, Wei, and
  Huang}]{ji2020generating}
Haozhe Ji, Pei Ke, Shaohan Huang, Furu Wei, and Minlie Huang.
  2020{\natexlab{a}}.
\newblock Generating commonsense explanation by extracting bridge concepts from
  reasoning paths.
\newblock \emph{arXiv preprint arXiv:2009.11753}.

\bibitem[{Ji et~al.(2020{\natexlab{b}})Ji, Ke, Huang, Wei, Zhu, and
  Huang}]{ji2020language}
Haozhe Ji, Pei Ke, Shaohan Huang, Furu Wei, Xiaoyan Zhu, and Minlie Huang.
  2020{\natexlab{b}}.
\newblock Language generation with multi-hop reasoning on commonsense knowledge
  graph.
\newblock \emph{arXiv preprint arXiv:2009.11692}.

\bibitem[{Khashabi et~al.(2020)Khashabi, Khot, Sabharwal, Tafjord, Clark, and
  Hajishirzi}]{khashabi2020unifiedqa}
Daniel Khashabi, Tushar Khot, Ashish Sabharwal, Oyvind Tafjord, Peter Clark,
  and Hannaneh Hajishirzi. 2020.
\newblock Unifiedqa: Crossing format boundaries with a single qa system.
\newblock \emph{arXiv preprint arXiv:2005.00700}.

\bibitem[{Kiesler(1983)}]{kiesler19831982}
Donald~J Kiesler. 1983.
\newblock The 1982 interpersonal circle: A taxonomy for complementarity in
  human transactions.
\newblock \emph{Psychological review}, 90(3):185.

\bibitem[{Levesque et~al.(2012)Levesque, Davis, and Morgenstern}]{Levesque2011}
Hector Levesque, Ernest Davis, and Leora Morgenstern. 2012.
\newblock The winograd schema challenge.
\newblock In \emph{Thirteenth International Conference on the Principles of
  Knowledge Representation and Reasoning}. Citeseer.

\bibitem[{Liu et~al.(2019)Liu, Ott, Goyal, Du, Joshi, Chen, Levy, Lewis,
  Zettlemoyer, and Stoyanov}]{liu2019roberta}
Yinhan Liu, Myle Ott, Naman Goyal, Jingfei Du, Mandar Joshi, Danqi Chen, Omer
  Levy, Mike Lewis, Luke Zettlemoyer, and Veselin Stoyanov. 2019.
\newblock Roberta: A robustly optimized bert pretraining approach.
\newblock \emph{arXiv preprint arXiv:1907.11692}.

\bibitem[{LoBue and Yates(2011)}]{LoBue2011}
Peter LoBue and Alexander Yates. 2011.
\newblock \href {http://www.cis.temple.edu/} {{Types of common-sense knowledge
  needed for recognizing textual entailment}}.
\newblock In \emph{ACL-HLT 2011 - Proceedings of the 49th Annual Meeting of the
  Association for Computational Linguistics: Human Language Technologies},
  volume~2, pages 329--334.

\bibitem[{Lourie et~al.(2021)Lourie, Bras, Bhagavatula, and
  Choi}]{lourie2021unicorn}
Nicholas Lourie, Ronan~Le Bras, Chandra Bhagavatula, and Yejin Choi. 2021.
\newblock Unicorn on rainbow: A universal commonsense reasoning model on a new
  multitask benchmark.
\newblock \emph{arXiv preprint arXiv:2103.13009}.

\bibitem[{Malaviya et~al.(2020)Malaviya, Bhagavatula, Bosselut, and
  Choi}]{Malaviya2019}
Chaitanya Malaviya, Chandra Bhagavatula, Antoine Bosselut, and Yejin Choi.
  2020.
\newblock Commonsense knowledge base completion with structural and semantic
  context.
\newblock In \emph{Proceedings of the AAAI Conference on Artificial
  Intelligence}, volume~34, pages 2925--2933.

\bibitem[{Mitra et~al.(2019)Mitra, Banerjee, Pal, Mishra, and
  Baral}]{mitra2019additional}
Arindam Mitra, Pratyay Banerjee, Kuntal~Kumar Pal, Swaroop Mishra, and Chitta
  Baral. 2019.
\newblock How additional knowledge can improve natural language commonsense
  question answering?
\newblock \emph{arXiv preprint arXiv:1909.08855}.

\bibitem[{Mostafazadeh et~al.(2016)Mostafazadeh, Chambers, He, Parikh, Batra,
  Vanderwende, Kohli, and Allen}]{mostafazadeh2016corpus}
Nasrin Mostafazadeh, Nathanael Chambers, Xiaodong He, Devi Parikh, Dhruv Batra,
  Lucy Vanderwende, Pushmeet Kohli, and James Allen. 2016.
\newblock A corpus and evaluation framework for deeper understanding of
  commonsense stories.
\newblock \emph{arXiv preprint arXiv:1604.01696}.

\bibitem[{Raffel et~al.(2019)Raffel, Shazeer, Roberts, Lee, Narang, Matena,
  Zhou, Li, and Liu}]{raffel2019exploring}
Colin Raffel, Noam Shazeer, Adam Roberts, Katherine Lee, Sharan Narang, Michael
  Matena, Yanqi Zhou, Wei Li, and Peter~J Liu. 2019.
\newblock Exploring the limits of transfer learning with a unified text-to-text
  transformer.
\newblock \emph{arXiv preprint arXiv:1910.10683}.

\bibitem[{Roemmele et~al.(2011)Roemmele, Bejan, and
  Gordon}]{roemmele2011choice}
Melissa Roemmele, Cosmin~Adrian Bejan, and Andrew~S Gordon. 2011.
\newblock Choice of plausible alternatives: An evaluation of commonsense causal
  reasoning.
\newblock In \emph{AAAI Spring Symposium: Logical Formalizations of Commonsense
  Reasoning}, pages 90--95.

\bibitem[{Sap et~al.(2019{\natexlab{a}})Sap, Le~Bras, Allaway, Bhagavatula,
  Lourie, Rashkin, Roof, Smith, and Choi}]{sap2019atomic}
Maarten Sap, Ronan Le~Bras, Emily Allaway, Chandra Bhagavatula, Nicholas
  Lourie, Hannah Rashkin, Brendan Roof, Noah~A Smith, and Yejin Choi.
  2019{\natexlab{a}}.
\newblock Atomic: An atlas of machine commonsense for if-then reasoning.
\newblock In \emph{Proceedings of the AAAI Conference on Artificial
  Intelligence}, volume~33, pages 3027--3035.

\bibitem[{Sap et~al.(2019{\natexlab{b}})Sap, Rashkin, Chen, LeBras, and
  Choi}]{sap2019socialiqa}
Maarten Sap, Hannah Rashkin, Derek Chen, Ronan LeBras, and Yejin Choi.
  2019{\natexlab{b}}.
\newblock Socialiqa: Commonsense reasoning about social interactions.
\newblock \emph{arXiv preprint arXiv:1904.09728}.

\bibitem[{Shen et~al.(2020)Shen, Mao, He, Long, Trischler, and
  Chen}]{shen2020exploiting}
Tao Shen, Yi~Mao, Pengcheng He, Guodong Long, Adam Trischler, and Weizhu Chen.
  2020.
\newblock Exploiting structured knowledge in text via graph-guided
  representation learning.
\newblock In \emph{Proceedings of the 2020 Conference on Empirical Methods in
  Natural Language Processing (EMNLP)}, pages 8980--8994.

\bibitem[{Shwartz et~al.(2020)Shwartz, West, Bras, Bhagavatula, and
  Choi}]{shwartz2020unsupervised}
Vered Shwartz, Peter West, Ronan~Le Bras, Chandra Bhagavatula, and Yejin Choi.
  2020.
\newblock Unsupervised commonsense question answering with self-talk.
\newblock \emph{arXiv preprint arXiv:2004.05483}.

\end{thebibliography}
\bibliographystyle{acl_natbib}

\clearpage
\appendix
\section{Appendix: Experiment Setting}
The number of hyperparameter search trials is 12 for each model. During training, each epoch takes about $30$ mins on average.
The hyperparameter configurations for best-performing models are listed as below.
\begin{table}[h]
\centering
\begin{adjustbox}{max width=0.45\textwidth}
\begin{tabular}{llll}
\toprule
Model                                                                                     &  \begin{tabular}[c]{@{}l@{}}Learning \\ Rate\end{tabular}            &  \begin{tabular}[c]{@{}l@{}}Batch \\ Size\end{tabular} & \begin{tabular}[c]{@{}l@{}}Gradient \\ Accumulation\end{tabular} \\
\midrule
RoBERTa Large                                                                             & $1e-5$ & 8          & 8                                                               \\
\midrule
+Question Relation                                                                        & $1e-5$ & 8          & 8                                                                \\
\midrule
+Social Knowkedge Cat                                                                & $1e-5$ & 8          & 4                                                              \\
\midrule
\begin{tabular}[c]{@{}l@{}}+Question Relation \\ + Social Knowledge Cat\end{tabular} & $1e-5$ & 8          & 16        \\                                                       \bottomrule
\end{tabular}
\end{adjustbox}
\end{table}

\section{Appendix: Case Study}
We conduct a case study to understand how the two types of categorical information help the QA task. The examples are listed in Table \ref{table:casestudy1}, \ref{table:casestudy2}.

\begin{table}[h]
\begin{adjustbox}{max width=0.45\textwidth}
\begin{tabular}{l}
\toprule
Question Relation Type: \textbf{xNeed}\\
\midrule
\makecell[l]{Riley told Austin's landlord that Austin was \\making a lot of noise at late hours. } \\
\midrule
What does Riley need to do before this?   \\
\midrule
\begin{tabular}[c]{@{}l@{}}(a) potentially call the police\\ (b) document incidents \Checkmark $\mid$ \textbf{Question Relation}\\ (c) follow up with the landlord $\mid$ \textbf{RoBERTa} \end{tabular}\\
\bottomrule
\end{tabular}
\end{adjustbox}
\caption{Case Study for Question Relation \\(\Checkmark: golden label)}
\label{table:casestudy1}
\end{table}

\begin{table}[h]
\centering
\begin{adjustbox}{max width=0.45\textwidth}
\begin{tabular}{l}
\toprule
\makecell[l]{Social Knowledge Category: \\ \textbf{Feelings and Characteristics}}\\
\midrule
\makecell[l]{Austin was taking a test and found it \\difficult at first.} \\
\midrule
How would you describe Austin?   \\
\midrule
\begin{tabular}[c]{@{}l@{}}(a) student $\mid$ \textbf{RoBERTa} \\ (b) stupid  \\ (c) overwhelmed \Checkmark $\mid$ \textbf{Social Knowledge Category} \end{tabular}\\
\bottomrule
\end{tabular}
\end{adjustbox}
\caption{Case Study for Social Knowledge Category (\Checkmark: golden label)}
\label{table:casestudy2}
\end{table}

\subsection{Question Relation Type}
Table \ref{table:casestudy1} refers to an example that is predicted correctly by the \textit{Question Relation} model but wrongly by the RoBERTa model. 
The RoBERTa model mistakenly selects the answer 
describing an event that could happen after the noise complaint. However, the question is asking possible events before it. The question type ``xNeed'' exactly provides such signal and indicates that the right answer should happen before the given context, and thus helps the QA model choose the right answer.

\subsection{Social Knowledge Category}
In Table \ref{table:casestudy2}, although the RoBERTa model selects the reasonable answer ``student'' to describe Austin, it fails to infer more in-depth semantic information embedded in the context. The actual focus is that Austin found the test challenging and was overwhelmed. Our social knowledge taxonomy assigns this example to the ``Feelings and Characteristics'' category. It enables the QA model to pay more attention to candidates that emphasize a kind of personal feeling.

\end{document}